\documentclass[review,12pt]{elsarticle}

\usepackage{graphicx}
\usepackage{amsmath,amssymb}
\usepackage{booktabs}
\usepackage{multirow}
\usepackage{algorithm}
\usepackage{algpseudocode}
\usepackage{hyperref}
\hypersetup{hidelinks,hypertexnames=false}
\biboptions{sort&compress}

\journal{Measurement / Advanced Engineering Informatics / Results in Engineering (target-dependent)}

\begin{document}

\begin{frontmatter}

\title{Leakage-Robust Evaluation and Data-Scale Sensitivity of Attention-Enhanced
Multi-Task Learning for Joint Fault Diagnosis and Remaining Useful Life Estimation}

\author[aff1]{Md Mahamudur Rahaman Shamim}
\ead{shamimwiu@gmail.com}

\author[aff1]{Md. Nuruzzaman}
\ead{nnuruzzaman1989@gmail.com}

\author[aff2]{Zannatul Ferdus\corref{cor2}}
\ead{zferdus508@gmail.com}
\cortext[cor2]{Corresponding author.}

\author[aff3]{Md Rajib Ahmed}
\ead{rajib.rahmed@gmail.com}

\author[aff4]{Abieer Nwshad Anwar}
\ead{abieernwshadanwar@gmail.com}

\author[aff5]{Mohammad Tooneer}
\ead{md.tooneer@gmail.com}

\author[aff6]{Johir Uddin Khan}
\ead{sajibjohir007@gmail.com}

\author[aff7]{Khalid Hossen\corref{cor1}}
\ead{khalid.hossen616@gmail.com}
\cortext[cor1]{Corresponding author.}

\address[aff1]{Department of Manufacturing Engineering Technology, Western Illinois University, Macomb, IL-61455, USA}
\address[aff2]{Department of Mechanical and Production Engineering, Ahsanullah University of Science and Technology, 141-142 Love Road, Tejgaon Industrial Area, Dhaka, Bangladesh}
\address[aff3]{Master of Science in Industrial/Engineering Management, Trine University, 720 Thunder Dr, Angola, IN 46703, USA}
\address[aff4]{Department of Nano-technology Engineering, Sapienza University of Rome, Rome, Italy}
\address[aff5]{Department of Industrial and Production Engineering, Bangladesh University of Engineering and Technology, Dhaka, Bangladesh}
\address[aff6]{Department of Mechanical Engineering, Chittagong University of Engineering and Technology, Raozan, Chattogram 4349, Bangladesh}
\address[aff7]{Department of Mechatronics Engineering, Rajshahi University of Engineering \& Technology (RUET), Kazla, Rajshahi, Bangladesh}

\begin{abstract}
Multi-task deep learning models that jointly perform fault classification and
remaining useful life (RUL) regression are increasingly proposed for
predictive maintenance (PdM), typically evaluated on sliding-window
segmentations of run-to-failure sensor data. We show that the train/test
splitting protocol applied to these windowed sequences has a first-order
effect on reported performance that is rarely audited in the PdM literature.
Using an attention-enhanced multi-task architecture (AMTLNet) evaluated on
three public benchmarks -- NASA C-MAPSS (turbofan degradation), NASA IMS
(bearing run-to-failure), and the UCI Hydraulic System dataset -- we
demonstrate empirically that naive splitting strategies can inflate
classification accuracy from a genuine $\sim$20--60\% to an artefactual
99.9\%, or conversely collapse it to 0\% through degenerate class
representation, depending on how sliding windows interact with the
train/test boundary. We introduce a chunk-based, leakage-audited splitting
protocol that eliminates both failure modes and report results under this
protocol across all three datasets with full seed-level statistics (5 seeds,
one-way ANOVA, Tukey HSD post-hoc). On the large-scale C-MAPSS benchmark
(19,976 leakage-free training windows), AMTLNet matches the strongest
single-task baseline (CNN-LSTM) on both classification (84.12$\pm$0.96\%
accuracy, Tukey $p=1.0$ vs.\ CNN-LSTM) and regression ($R^2=0.86\pm0.01$)
while significantly outperforming a naive multi-task baseline. On the two
small, single-run benchmarks (Bearing: $\sim$4,000 raw samples; Hydraulic:
$\sim$2,200 raw samples), multi-task training becomes unstable, and,
notably, instability manifests asymmetrically: the classification head is
the failure point for Bearing while the regression head is the failure
point for Hydraulic. We relate this asymmetry to the provenance of each
task's labels (measured vs.\ derived/proxy) and propose a practical decision
framework for when joint multi-task training is appropriate given available
sample volume. A controlled architectural ablation further shows that
AMTLNet's multi-head attention branch is the primary contributor to
regression stability (removing it alone drops $R^2$ from 0.861 to 0.766
and more than doubles classification variance across seeds), while the
convolutional branch contributes comparatively little to regression
quality despite accounting for roughly a third of the model's parameters.
This work contributes (i) an empirically demonstrated,
reusable leakage-audit protocol for windowed PdM benchmarks, (ii) a
statistically rigorous, seed-transparent evaluation of multi-task gains
that neither overstates nor hides negative results, and (iii) evidence that
task-specific stability under data scarcity depends on label provenance
rather than task type alone.
\end{abstract}



\end{frontmatter}

\section{Introduction}
\label{sec:intro}

Prognostics and health management (PHM) systems increasingly rely on deep
learning models trained on sliding-window segmentations of multivariate
sensor time series to jointly estimate a machine's current fault condition
and its remaining useful life (RUL)
\citep{qiu2023review,li2021perspective,sunal2022review}. Framing fault
diagnosis and RUL estimation as a single multi-task learning (MTL) problem,
rather than two independently trained models, is attractive: a shared
representation can, in principle, transfer degradation-relevant information
between the classification and regression objectives, reducing both
deployment cost and the risk of inconsistent predictions between the two
outputs \citep{liu2020jlcnn,behera2024ganmtl,abdelsamie2025mtlreview}.
Architectures combining convolutional feature extraction with recurrent or
attention-based temporal modelling -- CNN-LSTM, CNN-BiLSTM, and
attention-augmented variants -- have been reported to achieve strong
single-task RUL performance on standard benchmarks such as NASA C-MAPSS
\citep{deng2024cnnlstmattn,zhao2020doublechannel,ghoneim2025attention,han2024dualattention,wu2025palcnet},
and multi-task extensions have been proposed for bearing and hydraulic
system monitoring \citep{kim2025semisupervised,bisoyi2025ecnngrusam,wang2026msranate,jafari2026hydraulic,kim2022hydraulic}.

However, a methodological question that is rarely addressed explicitly in
this literature is how the sliding windows used to segment run-to-failure
sequences interact with the train/validation/test split. Because
consecutive windows generated with a small stride overlap substantially --
often sharing all but one timestep with their neighbour -- a train/test
split performed at the level of individual windows, rather than at the
level of contiguous time segments, can place near-duplicate windows on
both sides of the split. This is a specific instance of a broader,
well-documented problem in time-series and sequential machine learning:
data leakage arising from improper partitioning inflates reported
performance and undermines the validity of benchmark comparisons
\citep{kapoor2023leakage,sasse2023leakagescenarios,geirhos2020shortcut}.
The effect has been quantified directly in general time-series and
sequence-classification settings
\citep{tampu2022octleakage,rosenblatt2024connectome,yang2024emd}, and
closely related failure modes -- single random splits producing unreliable
estimates on small samples, and unaccounted seed-to-seed variance
inflating or obscuring true effect sizes -- have been documented across
biomedical imaging, causal inference, and general machine-learning
benchmarking
\citep{an2021radiomics,schader2024seed,maleki2022generalizability,calle2025nestedcv,banerjee2024variability}.
Within the PdM literature specifically, however, published sliding-window
pipelines for C-MAPSS, IMS Bearing, and hydraulic-system RUL estimation
typically report a single train/test split with stride-1 windowing and do
not report a leakage audit of the kind now considered standard practice in
adjacent fields \citep{hewamalage2022forecast,lema2025benchmarking}.

A second, related gap concerns the data-volume assumptions implicit in
applying deep multi-task architectures to PdM benchmarks. NASA C-MAPSS
provides tens of thousands of samples from 100 simulated engine units,
but many other widely used PdM benchmarks -- including the NASA IMS
bearing dataset and the UCI Hydraulic System dataset used in this study --
consist of a small number of continuous run-to-failure or condition-labelled
trials, yielding at most a few thousand raw samples. The small-data
literature in PHM has extensively documented that deep learning
performance degrades under such data scarcity, motivating few-shot,
meta-learning, and transfer-learning remedies
\citep{li2024smalldata,wang2021metalearning,zhang2019limiteddata,wu2020fewshot,liang2023fewshotreview}.
What remains underexplored is how \emph{multi-task} training specifically
behaves as data volume shrinks, relative to single-task training on the
same data -- whether both tasks degrade in tandem, whether one task
protects the other through shared representation learning, or whether
degradation is asymmetric and, if so, what predicts which task will be
more fragile. Negative transfer between jointly trained tasks has been
studied in general multi-task learning
\citep{huang2022curriculum,depaula2026negativetransfer,wang2023hardsharing},
but rarely with an explicit link to the provenance of each task's ground
truth (directly measured vs.\ derived/proxy labels), which we show below is
a strong predictor of task-specific fragility in our benchmarks.

Building on co-author Hossen's prior work establishing leakage-free windowing and
provenance-aware data partitioning as a methodological requirement for
deployable vibration-based diagnostics \citep{hossen2026domainshift}, and on
his earlier comparative evaluation of classical machine learning algorithms
for vibration-based fault diagnosis \citep{hossen2025compressor}, this paper
makes the following contributions:

\begin{enumerate}
\item We empirically demonstrate, on the same architecture and the same
raw data, how three distinct sliding-window splitting protocols produce
qualitatively different and in two cases clearly invalid performance
estimates (Section~\ref{sec:protocol-ablation}), and we introduce a
chunk-based, leakage-audited protocol that avoids all three failure
modes.
\item We report a statistically rigorous evaluation of an
attention-enhanced multi-task architecture (AMTLNet) against single-task
and naive multi-task baselines on NASA C-MAPSS under this leakage-audited
protocol, including one-way ANOVA and Tukey HSD post-hoc testing across
five random seeds, and show that AMTLNet matches the best single-task
baseline while significantly outperforming naive multi-task training.
\item We characterise, with full seed-level transparency, how the same
multi-task architecture destabilises on two small, single-run PdM
benchmarks, and show that the destabilisation is task-asymmetric and
correlates with whether each task's labels are directly measured or
derived proxies.
\item We translate these findings into a practical decision framework for
practitioners choosing between joint multi-task and single-task
deployment given the sample volume available for a given PdM asset.
\end{enumerate}

The remainder of this paper is organised as follows. Section~\ref{sec:related}
reviews related work in RUL/fault-diagnosis deep learning, multi-task PdM,
and data leakage in sequential machine learning. Section~\ref{sec:methods}
describes the datasets, the leakage-audited splitting protocol, the AMTLNet
architecture, and the statistical evaluation procedure.
Section~\ref{sec:results} presents the protocol ablation and the main
results. Section~\ref{sec:discussion} discusses the data-scale sensitivity
finding, its practical implications, and the study's limitations.
Section~\ref{sec:conclusion} concludes.

\section{Related Work}
\label{sec:related}

\subsection{Deep learning for RUL prediction and fault diagnosis}
CNN, LSTM, and hybrid CNN-LSTM/BiLSTM architectures are the dominant
approach to sliding-window RUL estimation on NASA C-MAPSS
\citep{zhao2020doublechannel,deng2024cnnlstmattn,borst2023cnnlstm,han2024dualattention},
with attention mechanisms increasingly used to weight informative sensors
or timesteps \citep{ghoneim2025attention,wu2025palcnet,fan2025transformer}.
Parallel work applies the same architectural family to bearing
run-to-failure data \citep{kim2025semisupervised,bisoyi2025ecnngrusam,zhang2019limiteddata}
and to hydraulic and pump condition monitoring
\citep{kim2022hydraulic,jafari2026hydraulic,wang2026msranate,sunal2022review}.
Comprehensive surveys document the breadth of this literature and its
continuing reliance on deep feature extraction under matched
training/deployment conditions \citep{qiu2023review,li2021perspective,saeed2025resource}.

\subsection{Multi-task learning for predictive maintenance}
Jointly training fault classification and RUL regression heads on a shared
encoder has been proposed to exploit correlated degradation information
between the two tasks \citep{liu2020jlcnn,behera2024ganmtl}. Broader
surveys of multi-task learning describe hard- and soft-parameter-sharing
architectures and their trade-offs \citep{abdelsamie2025mtlreview,wang2023hardsharing},
and negative transfer -- where jointly training tasks harms one or both
relative to single-task training -- is a recognised risk, studied both in
reinforcement learning and language-processing contexts
\citep{huang2022curriculum,depaula2026negativetransfer}. However, few
studies systematically compare multi-task against single-task performance
\emph{as a function of dataset scale} within a single controlled PdM study,
which is the focus of Section~\ref{sec:results} of this paper.

\subsection{Data leakage and evaluation rigor in sequential machine learning}
Data leakage -- the unintentional exposure of test-set information during
training -- is increasingly recognised as a systemic issue undermining
reported performance across machine-learning-based science
\citep{kapoor2023leakage,sasse2023leakagescenarios}. In time-series and
sequential settings specifically, leakage frequently arises when
overlapping sliding windows are generated prior to, rather than after,
train/test partitioning, allowing near-duplicate segments to appear on
both sides of the split \citep{yang2024emd,hewamalage2022forecast}. This
effect has been directly quantified in biomedical imaging
\citep{tampu2022octleakage}, neuroimaging \citep{rosenblatt2024connectome},
recommender systems \citep{ji2020recsys}, and large language model
benchmarking \citep{lopez2024llmleakage}, and is closely related to the
broader phenomenon of shortcut learning, in which models exploit spurious,
non-generalisable correlations available in a particular evaluation setup
rather than learning the intended task \citep{geirhos2020shortcut}. A
distinct but related failure mode -- unreliable performance estimates from
a single random split on small samples, and unreported seed-to-seed
variance -- has been documented in radiomics
\citep{an2021radiomics}, causal inference \citep{schader2024seed}, and
general deep-learning benchmarking
\citep{maleki2022generalizability,calle2025nestedcv,banerjee2024variability},
motivating the multi-seed, statistically tested evaluation protocol we
adopt in this work. Within PdM specifically, reproducible and
statistically rigorous benchmarking practices have been proposed for
surface-defect detection \citep{lema2025benchmarking} and synthetic
building-analytics data \citep{miller2025smartbuildsim}, but, to our
knowledge, an explicit leakage audit of sliding-window splitting has not
previously been reported for the C-MAPSS/IMS Bearing/Hydraulic benchmark
family jointly used in multi-task PdM studies.

\subsection{Small-data challenges in prognostics and health management}
Data scarcity is a well-recognised obstacle to deploying deep learning in
industrial PHM, where high-cost data collection limits the volume of
labelled run-to-failure data available for a given asset class
\citep{li2024smalldata}. Few-shot and meta-learning approaches have been
proposed specifically for bearing fault diagnosis under limited samples
\citep{wang2021metalearning,zhang2019limiteddata,wu2020fewshot,liang2023fewshotreview},
and transfer learning is widely used to compensate for insufficient
target-domain data \citep{han2018deeptransfer,zhang2024diesel}. Our own
prior work shows that even a fixed, compact architecture exhibits strongly
domain-dependent generalisation when transferred across vibration sensing
conditions, underscoring that data volume and data provenance jointly
determine achievable robustness rather than either factor alone
\citep{hossen2026domainshift}. This paper extends that line of inquiry from
single-task cross-domain transfer to multi-task training stability as a
function of within-domain sample volume.

\section{Materials and Methods}
\label{sec:methods}

\subsection{Datasets}
\label{sec:datasets}
Three public benchmarks spanning three orders of magnitude in raw sample
volume are used, selected to allow a controlled comparison of multi-task
training stability across dataset scale.

\textbf{NASA C-MAPSS (FD001).} Simulated turbofan engine degradation data
comprising 100 engine units, 20{,}631 raw time-cycle rows, and 21 sensor
channels (16 retained after removing near-zero-variance sensors). Fault
condition is discretised into four ordinal classes from run-to-failure
position, and RUL is computed per-unit with a standard piecewise cap at
125 cycles.

\textbf{NASA IMS Bearing.} Vibration data from a run-to-failure bearing
test rig (2nd\_test subset), subsampled to 200{,}000 raw rows across four
channels and reduced to 16 engineered statistical features (RMS, standard
deviation, kurtosis, peak amplitude per channel) via a 50-sample rolling
window, yielding 3{,}999 feature rows. Fault condition is discretised into
three classes (rebalanced from an initial four-class scheme; see
Section~\ref{sec:protocol-ablation}) based on position within the
degradation trajectory.

\textbf{UCI Hydraulic System.} Condition-monitoring data from a hydraulic
test rig, comprising 2{,}205 cycles with seven low-frequency sensor
channels (temperature, vibration, cooling efficiency/power). Fault
condition corresponds to the dataset's three directly measured cooler
condition labels (3, 20, 100); RUL is a synthetic proxy constructed as a
weighted combination of cooler, valve, pump, and accumulator condition
codes.

We deliberately retain the label-construction differences between
datasets -- C-MAPSS's cycle-based RUL, Bearing's position-derived class
and RUL proxy, and Hydraulic's directly measured class but synthetic RUL
proxy -- because, as shown in Section~\ref{sec:results}, this provenance
difference is central to explaining the asymmetric instability we observe.
Table~\ref{tab:dataset-summary} summarises the three datasets at a
glance, making explicit that this study deliberately spans three
different data scales and three different label-provenance regimes.

\begin{table}[htbp]
\centering
\caption{Dataset summary: scale and label provenance.}
\label{tab:dataset-summary}
\footnotesize
\resizebox{\textwidth}{!}{%
\begin{tabular}{lllll}
\toprule
Dataset & Raw samples & Features & Class label & RUL label \\
\midrule
C-MAPSS FD001 & 20{,}631 rows & 16 (14 sensor + 2 setting) & RUL-threshold, 4 class, derived & capped @125, derived \\
IMS Bearing   & 3{,}999 feature rows & 16 statistical & trajectory-position, 3 class, derived/proxy & linear interp., derived/proxy \\
UCI Hydraulic & 2{,}205 cycles & 13 sensor (26 stat.\ features) & cooler code, 3 class, \textbf{measured} & weighted proxy, derived \\
\bottomrule
\end{tabular}%
}
\end{table}

\subsection{Preprocessing and label construction}
\label{sec:preprocessing}
To make every reported class and RUL value fully reproducible, we state
the exact construction rule used for each dataset.

\textbf{C-MAPSS.} Of the 21 raw sensor channels, 7 are dropped for
near-zero variance (a well-known subset in the literature), leaving 14
sensor channels plus 2 operational-setting channels (16 input features
total). RUL is computed per engine unit as
$\mathrm{RUL} = \min(\mathrm{cycle}_{\max} - \mathrm{cycle},\ 125)$,
i.e.\ the standard piecewise linear cap used throughout the RUL
literature. The four fault classes are thresholds on this RUL value:
class 0 (Healthy) for $\mathrm{RUL} > 80$, class 1 (Early degradation)
for $40 < \mathrm{RUL} \le 80$, class 2 (Advanced) for
$10 < \mathrm{RUL} \le 40$, and class 3 (Critical) for
$\mathrm{RUL} \le 10$.

\textbf{IMS Bearing.} The four raw vibration channels (200{,}000
subsampled rows) are reduced via a non-overlapping 50-sample rolling
window into 16 statistical features (RMS, standard deviation, kurtosis,
and peak amplitude, per channel), yielding 3{,}999 feature rows. Because
no physically measured fault-onset label exists for this dataset, RUL is
constructed as a linear interpolation over trajectory position,
$\mathrm{RUL}_i = 125 \times (1 - i/n)$ for feature-row index $i$ of $n$
total rows, and the fault class is a threshold on the corresponding
progress fraction $p_i = i/n$: class 0 for $p_i < 0.55$, class 1 for
$0.55 \le p_i < 0.80$, and class 2 for $p_i \ge 0.80$ (rebalanced from an
initial four-class, 60/80/95 threshold scheme whose top bucket contained
too few samples to train reliably; see Section~\ref{sec:protocol-ablation}).
We flag this construction explicitly: both the class label and the RUL
target are \emph{derived proxies} based on position within the
run-to-failure trajectory, not independently measured degradation
indicators.

\textbf{UCI Hydraulic.} Thirteen low-frequency sensor channels (pressure,
temperature, vibration, cooling efficiency and power) are reduced to 26
statistical features (mean and standard deviation per channel) per
2{,}205-cycle experiment. The fault class is the dataset's own directly
measured cooler-condition code (3, 20, or 100) -- a genuine physical
measurement, not a derived proxy. RUL, by contrast, has no directly
measured counterpart in this dataset and is constructed as a weighted
combination of all four condition variables:
\begin{equation}
\mathrm{RUL} = 0.4\, c_{\mathrm{cooler}} + 0.3\, (125\, v_{\mathrm{valve}}) + 0.2\, (125\, p_{\mathrm{pump}}) + 0.1\, (125\, a_{\mathrm{accum}})
\label{eq:hydraulic-rul}
\end{equation}
where $c_{\mathrm{cooler}} \in \{125, 62, 10\}$ maps the cooler code
$\{100, 20, 3\}$ to an RUL-scale value, $v_{\mathrm{valve}} =
(\mathrm{valve} - 73)/27$, $p_{\mathrm{pump}} = 1 - \mathrm{pump}/2$, and
$a_{\mathrm{accum}} = (\mathrm{accum} - 90)/40$ are min-max-style
normalisations of the raw valve, pump, and accumulator condition codes.
The 0.4/0.3/0.2/0.1 weighting reflects the cooler condition's dominant
influence on system RUL relative to the other three subsystems, based on
domain judgement rather than a measured ground truth; we treat this
explicitly as a modelling choice, not a physical certainty, throughout
the paper's discussion of this dataset's regression instability.

Making the exact class and RUL construction transparent here directly
addresses three questions a careful reader might otherwise be left to
guess at: how the fault classes were constructed, whether the Hydraulic
RUL proxy is physically meaningful, and whether Bearing's classification
instability (Section~\ref{sec:results}) reflects a genuine data
limitation or an artefact of label construction -- we return to this
last question directly in Section~\ref{sec:discussion}.

\subsection{Leakage-audited windowing and splitting protocol}
\label{sec:protocol}

\begin{figure}[htbp]
\centering
\includegraphics[width= 0.7\textwidth]{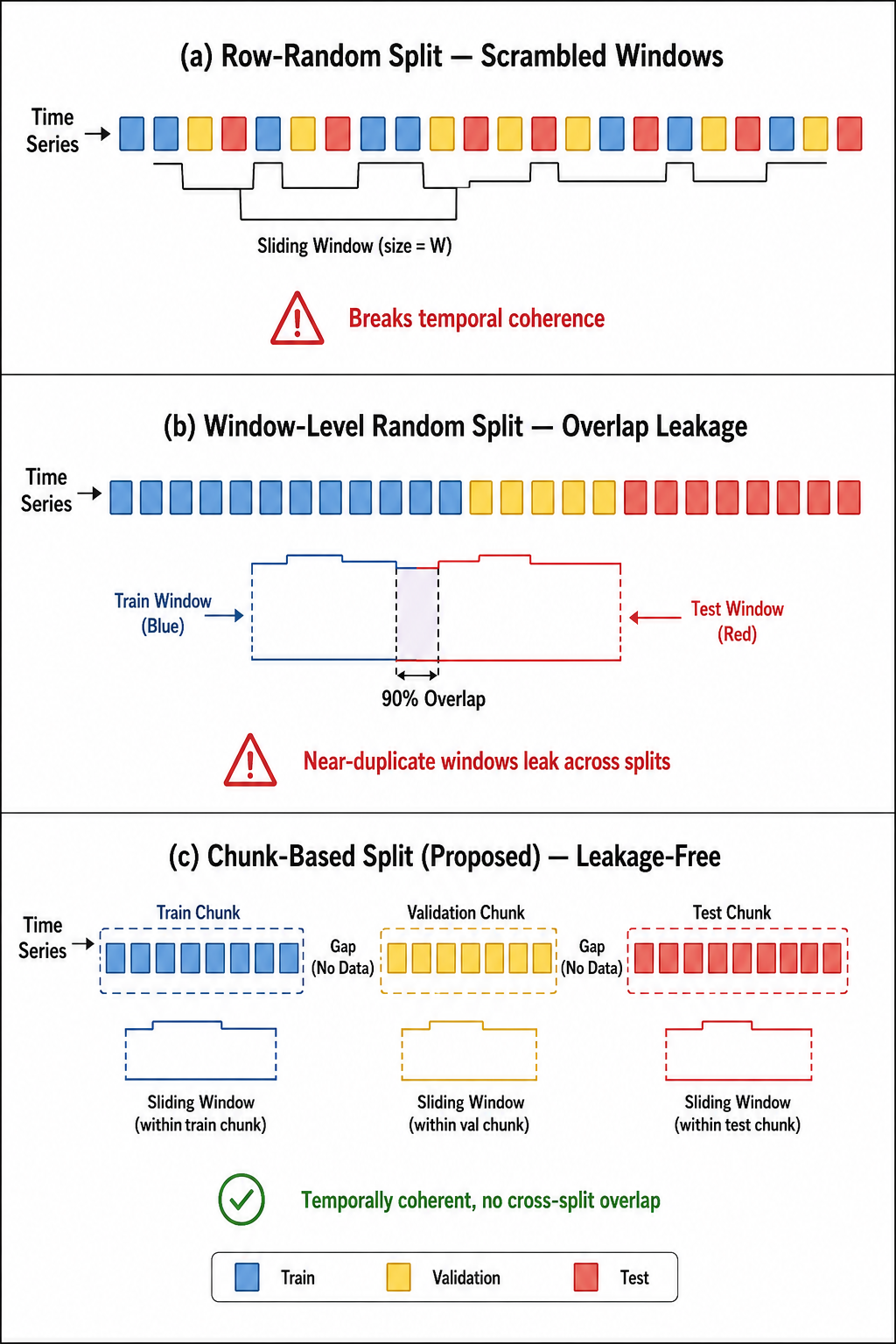}
\caption{Three sliding-window splitting strategies for run-to-failure
sensor data. (a) Row-level random splitting scrambles temporal order,
destroying window coherence. (b) Window-level random splitting under a
small stride places near-duplicate, overlapping windows across the
train/test boundary. (c) The proposed chunk-based protocol partitions the
raw sequence into contiguous, non-overlapping chunks prior to windowing,
eliminating both failure modes.}
\label{fig:leakage}
\end{figure}

Figure~\ref{fig:leakage} illustrates the three splitting strategies
compared empirically in Section~\ref{sec:protocol-ablation}: naive
row-level random splitting, which scrambles temporal order within a
window; naive window-level random splitting under a small stride, which
places near-duplicate overlapping windows on both sides of the split; and
the chunk-based protocol adopted in this work, which eliminates both
failure modes by splitting contiguous, non-overlapping chunks of the raw
signal before any windowing occurs.

All three datasets are converted to fixed-length sliding windows for input
to the sequence models. For a raw sequence of length $N$ with window size
$w$ and stride $s$, window $i$ spans rows $[i, i+w)$ and is labelled using
the class and RUL value at its final timestep, following standard
practice \citep{zhao2020doublechannel}. We do not window the raw sequence
in isolation; instead, we adopt a \textbf{chunk-based splitting protocol}
motivated directly by the empirical failure modes documented in
Section~\ref{sec:protocol-ablation}:

\begin{enumerate}
\item The raw, time-ordered sequence is partitioned into $n$ contiguous,
non-overlapping chunks, each spanning $w \times m$ rows for a
chunk-size multiplier $m$ (chunk counts and sizes for each dataset are
reported alongside the corresponding results in Section~\ref{sec:protocol-ablation}).
\item Each chunk is assigned a single dominant class label (its majority
class), and whole chunks -- not individual windows -- are randomly
assigned to train (70\%), validation (15\%), and test (15\%) splits,
stratified by dominant class.
\item Sliding windows are generated \emph{independently within each
chunk}, so that no window can span a chunk boundary and no two windows
assigned to different splits can share more than $w-1$ raw rows unless
they originate from the same chunk (in which case they are, by
construction, assigned to the same split).
\item Feature scaling (min-max normalisation) and RUL scaling (robust
scaling) are fit on the training split only and applied to validation and
test splits without refitting.
\item The training split (only) is, if necessary, capped to a maximum
window budget via stratified subsampling, to bound training cost on the
largest dataset (C-MAPSS).
\item Class balance in the (capped) training split is then addressed via
SMOTE oversampling, applied to the training split only; validation and
test splits are never resampled, so reported performance always reflects
the natural, unmodified class distribution.
\end{enumerate}

This protocol guarantees that (i) every window is constructed from a
temporally contiguous segment of the original signal, and (ii) no window
in the test set can share more than a negligible fraction of raw
timesteps with any window in the training set. Class-weighted training
was evaluated as an additional stabilisation measure (Section~\ref{sec:protocol-ablation})
but found unnecessary once the splitting protocol above was in place,
since class-weighted training loss did not exhibit measurably imbalanced
weights under this protocol's stratified chunk assignment. Algorithm~\ref{alg:chunksplit}
summarises the procedure formally.

\begin{algorithm}[htbp]
\caption{Chunk-based leakage-audited split}
\label{alg:chunksplit}
\begin{algorithmic}[1]
\State \textbf{Input:} raw sequence $X, y_{\mathrm{cls}}, y_{\mathrm{rul}}$ (sorted by unit and time), window size $w$, stride $s$, chunk multiplier $m$
\State Set chunk size $c \gets \max(w \times m,\ w+1)$
\State Partition $X$ into contiguous, non-overlapping chunks of size $\approx c$
\State Assign each chunk its majority class label from $y_{\mathrm{cls}}$
\State Stratify whole chunks (not windows) into Train / Val / Test (70/15/15) by majority class
\For{each split $\in \{$Train, Val, Test$\}$}
    \For{each chunk assigned to split}
        \State Generate sliding windows of size $w$, stride $s$, \emph{only within this chunk}
    \EndFor
\EndFor
\State Fit feature scaler and RUL scaler on Train windows only
\State Apply fitted scalers to Val and Test windows (no refitting)
\State \textbf{Audit:} verify zero raw-row overlap between windows in different splits \Comment{guaranteed by construction, Step 3}
\State \textbf{Output:} leakage-free Train / Val / Test window sets
\end{algorithmic}
\end{algorithm}

Because windows are only ever constructed within a single chunk, and
every chunk is assigned in its entirety to exactly one split, the
audited cross-split raw-row overlap is exactly zero by construction for
all three datasets under this protocol -- a guarantee rather than an
empirical measurement. Table~\ref{tab:protocol-params} reports the
concrete window, stride, and chunking parameters used per dataset,
together with the resulting split sizes both immediately after windowing
and after the class-balancing step (capping and/or SMOTE, applied to the
training split only) described in Step 5--6 of Section~\ref{sec:protocol}.

\begin{table}[htbp]
\centering
\caption{Splitting protocol parameters and resulting split sizes per
dataset. ``Train (windowed)'' is the training split size immediately
after chunk-based windowing; ``Train (final)'' is the training set size
after the capping and/or SMOTE class-balancing applied only to the
training split (Section~\ref{sec:protocol}), i.e.\ the number of samples
actually used to train each model. Validation and test sizes are
unaffected by this step and are reported once.}
\label{tab:protocol-params}
\footnotesize
\resizebox{\textwidth}{!}{%
\begin{tabular}{llllllll}
\toprule
Dataset & Window $w$ & Stride $s$ & Chunk mult.\ $m$ & \#Chunks & Train (windowed) & Train (final) & Val / Test \\
\midrule
C-MAPSS   & 30 & 1 & 5 & 137 & 11{,}400 & 19{,}976 (post-SMOTE) & 2{,}601 / 2{,}520 \\
Bearing   & 30 & 5 & 3 & 44  & 368      & 585 (post-SMOTE)      & 84 / 84 \\
Hydraulic & 10 & 2 & 5 & 44  & 600      & 600 (no SMOTE applied) & 140 / 143 \\
\bottomrule
\end{tabular}%
}
\end{table}

\subsection{AMTLNet architecture}
\label{sec:architecture}
\begin{figure}[htbp]
\centering
\includegraphics[width=\textwidth]{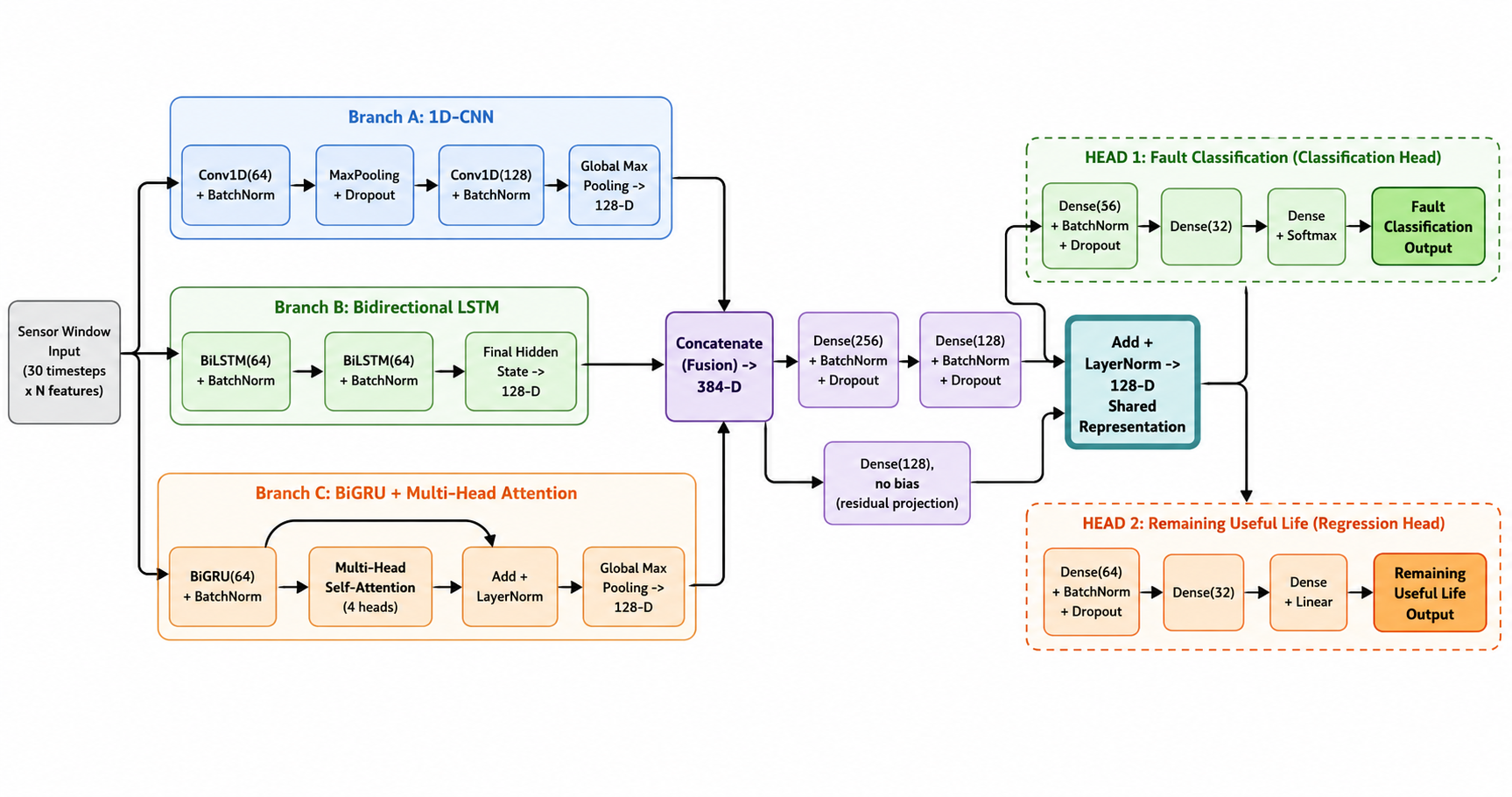}
\caption{AMTLNet architecture. A three-branch shared encoder (1D-CNN,
bidirectional LSTM, and bidirectional GRU with multi-head self-attention)
is concatenated and fused through a residual dense block into a shared
128-dimensional representation, from which a fault-classification head
and an RUL-regression head branch independently.}
\label{fig:architecture}
\end{figure}

Figure~\ref{fig:architecture} shows the overall architecture. Following our earlier design of a multi-task predictive-maintenance
architecture and drawing on established CNN/recurrent/attention hybrid
designs for sensor time series \citep{ghoneim2025attention,han2024dualattention},
the shared encoder is a \emph{three-branch} design. Branch A applies a 1D
convolutional stack (Conv1D--BatchNorm--MaxPooling--Dropout--Conv1D--BatchNorm)
followed by global max pooling to extract local temporal patterns (128-D).
Branch B applies a two-layer bidirectional LSTM
(BiLSTM--BatchNorm--BiLSTM--BatchNorm) to capture longer-range sequential
dependencies (128-D). Branch C applies a bidirectional GRU followed by a
multi-head self-attention block (4 heads, key dimension 32) with a
residual connection and layer normalisation, then global max pooling
(128-D). The three branch outputs are concatenated into a 384-dimensional
fused representation, passed through a two-layer dense block
(Dense(256)--BatchNorm--Dropout--Dense(128)--BatchNorm--Dropout) with a
parallel linear residual projection of the fused representation, summed
and layer-normalised to produce the final 128-dimensional shared
representation. From this representation, two task-specific heads
branch: a fault-classification head (Dense--Dense--Softmax) and an
RUL-regression head (Dense--Dense--Linear). Section~\ref{sec:futurework}
reports a controlled ablation isolating each of the three branches'
individual contribution, including a dedicated attention-ablation variant
in which the multi-head attention step is removed from Branch C while
retaining its bidirectional GRU encoding, isolating the attention
mechanism's specific contribution independent of branch removal. The
model is trained with a joint loss combining sparse categorical
cross-entropy for classification and Huber loss for regression, with
task loss weights of 2.0 and 1.5 respectively, using the Adam optimiser
with early stopping on validation RUL MAE (patience 25 epochs, maximum 80
epochs, batch size 256). The total parameter count is approximately
472{,}000.

Formally, writing $f_A(\cdot)$, $f_B(\cdot)$, $f_C(\cdot)$ for the three
branch encoders operating on an input window $X \in \mathbb{R}^{w \times
d}$, the fused representation is
\begin{equation}
h_{\mathrm{fuse}} = \big[\, f_A(X) \,\Vert\, f_B(X) \,\Vert\, f_C(X) \,\big] \in \mathbb{R}^{384}
\end{equation}
where $\Vert$ denotes concatenation. The shared representation is then
computed with a residual dense block,
\begin{equation}
h_{\mathrm{shared}} = \mathrm{LayerNorm}\big(\, \mathrm{MLP}(h_{\mathrm{fuse}}) + W_r\, h_{\mathrm{fuse}} \,\big) \in \mathbb{R}^{128}
\end{equation}
where $\mathrm{MLP}(\cdot)$ is the two-layer Dense(256)$\to$Dense(128)
block and $W_r \in \mathbb{R}^{128 \times 384}$ is the bias-free linear
residual projection. The joint training objective is
\begin{equation}
\mathcal{L} = w_{\mathrm{cls}} \cdot \mathcal{L}_{\mathrm{CE}}(y_{\mathrm{cls}}, \hat{y}_{\mathrm{cls}}) + w_{\mathrm{rul}} \cdot \mathcal{L}_{\mathrm{Huber}}(y_{\mathrm{rul}}, \hat{y}_{\mathrm{rul}})
\label{eq:jointloss}
\end{equation}
with $w_{\mathrm{cls}} = 2.0$ and $w_{\mathrm{rul}} = 1.5$. This weighting
was chosen to counteract the classification loss's typically smaller
numerical magnitude relative to Huber-loss regression residuals on
RobustScaler-normalised RUL targets, ensuring neither task's gradient
dominates training by default; Section~\ref{sec:futurework} identifies a
systematic loss-weight sensitivity analysis as a remaining extension
rather than a claim that this specific ratio is optimal.

We note explicitly, rather than leave implicit, that early stopping and
checkpoint selection (patience 25 epochs) monitor validation RUL MAE
only. This was a deliberate simplification -- a single scalar monitoring
criterion applicable across all three datasets and all baseline models
without per-dataset tuning -- rather than an oversight; however, as
Section~\ref{sec:limitations} discusses in detail, this criterion
provides no explicit safeguard against classification-head overfitting,
which we identify as a contributing factor to the Bearing classification
instability reported in Section~\ref{sec:results}.

\subsection{Baselines}
AMTLNet is compared against four baselines trained under the identical
leakage-audited protocol and identical windowed inputs: a single-task 1D-CNN
(CNN-STL), a single-task LSTM (LSTM-STL), a single-task CNN-LSTM hybrid
(CNN-LSTM-STL), and a naive multi-task LSTM without AMTLNet's three-branch
encoder design (Vanilla MTL-LSTM). All baselines share the same training budget,
optimiser, and early-stopping criterion as AMTLNet to isolate the effect
of architectural design from training-procedure differences.
Table~\ref{tab:baseline-summary} summarises each model's task type,
encoder, and role in the comparison.

\begin{table}[htbp]
\centering
\caption{Baseline model summary.}
\label{tab:baseline-summary}
\footnotesize
\resizebox{\textwidth}{!}{%
\begin{tabular}{llllll}
\toprule
Model & Task type & Encoder & Output heads & Role \\
\midrule
CNN-STL          & Single-task & Conv1D                  & Separate & Local-feature baseline \\
LSTM-STL         & Single-task & LSTM                    & Separate & Recurrent baseline \\
CNN-LSTM-STL     & Single-task & CNN + LSTM              & Separate & Strongest single-task baseline \\
Vanilla MTL-LSTM & Multi-task  & LSTM (shared)           & Joint    & Naive multi-task baseline \\
\textbf{AMTLNet} & Multi-task  & CNN + BiLSTM + BiGRU-attn (shared) & Joint & Proposed \\
\bottomrule
\end{tabular}%
}
\end{table}

\subsection{Training configuration}
\label{sec:trainconfig}
Table~\ref{tab:trainconfig} reports the complete training configuration
used for all models (AMTLNet, all baselines, and all ablation variants)
to support exact reproduction.

\begin{table}[htbp]
\centering
\caption{Training configuration.}
\label{tab:trainconfig}
\footnotesize
\begin{tabular}{ll}
\toprule
Item & Setting \\
\midrule
Framework                 & TensorFlow / Keras 3, GPU-accelerated \\
Optimiser                 & Adam \\
Learning rate              & $1\times10^{-3}$ (initial) \\
Batch size                & 256 \\
Maximum epochs            & 80 (main experiments); 40 (ablation study, Section~\ref{sec:ablation-results}) \\
Early stopping            & Patience 25 epochs, monitor: validation RUL MAE \\
Classification loss       & Sparse categorical cross-entropy \\
RUL loss                  & Huber loss ($\delta=1.0$) \\
Loss weights ($w_{\mathrm{cls}}, w_{\mathrm{rul}}$) & 2.0, 1.5 \\
Feature scaling           & MinMaxScaler, fit on training windows only \\
RUL scaling               & RobustScaler, fit on training windows only \\
Seeds                     & 42, 123, 2024, 7, 99 (5 seeds; 3 for ablation study) \\
Hardware                  & NVIDIA P100 GPU (Kaggle) \\
\bottomrule
\end{tabular}
\end{table}

\subsection{Statistical evaluation procedure}
\label{sec:stats}
Every architecture is trained from five independent random seeds
(42, 123, 2024, 7, 99) with the data split held fixed across seeds, so
that reported variance reflects training stochasticity (weight
initialisation, batch ordering) rather than data-split variability.
Five seeds were used to quantify training stochasticity while keeping
the data split fixed, so that reported variance reflects optimisation
and initialisation effects rather than split variation; we report this
choice explicitly as a practical budget rather than a claim of
statistical saturation, and note ten or more seeds would further tighten
confidence intervals at proportionally higher compute cost. We
report mean $\pm$ standard deviation across seeds for accuracy,
precision, recall, F1-score, MAE, RMSE, and $R^2$, and additionally
report the 95\% confidence interval on the mean,
\begin{equation}
\mathrm{CI}_{95\%} = \bar{x} \pm t_{0.975, n-1} \frac{s}{\sqrt{n}}
\label{eq:ci}
\end{equation}
with $n=5$ and $t_{0.975,4} = 2.776$, for the C-MAPSS headline metrics
(Section~\ref{sec:results}) to make the uncertainty of the mean estimate
explicit alongside seed-to-seed variability, which prevents overclaiming
precision from a 5-seed budget. For the C-MAPSS
comparison against baselines, we additionally report a one-way ANOVA on
seed-level F1-scores across all five models, followed by Tukey's HSD
post-hoc test with family-wise error rate control at $\alpha=0.05$, to
determine which pairwise differences are statistically supported rather
than relying on mean differences alone -- directly following recommended
practice for benchmarking under training stochasticity
\citep{banerjee2024variability,calle2025nestedcv}. We further report the
ANOVA effect size,
\begin{equation}
\eta^2 = \frac{SS_{\mathrm{between}}}{SS_{\mathrm{total}}}
\label{eq:eta2}
\end{equation}
computed from the ANOVA's $F$-statistic and degrees of freedom as
$\eta^2 = (F \cdot df_{\mathrm{between}}) / (F \cdot df_{\mathrm{between}} + df_{\mathrm{within}})$,
to quantify how much of the total variance in seed-level F1-scores is
attributable to model choice, independent of statistical significance.

\section{Results}
\label{sec:results}

\subsection{Protocol ablation: quantifying the cost of naive splitting}
\label{sec:protocol-ablation}
To isolate the effect of the splitting protocol from all other pipeline
choices, we trained the identical AMTLNet architecture on the identical
raw data under three splitting strategies, changing only the split logic
described in Section~\ref{sec:protocol}.

\textbf{Protocol A -- row-level random split, then windowing.} Raw rows
are randomly assigned to train/validation/test \emph{before} sliding
windows are constructed. Because standard train/test splitting functions
do not preserve temporal order among the selected rows, sliding windows
built from a temporally scrambled row order combine unrelated,
non-contiguous timesteps into a single ``sequence''. Under this protocol,
NASA IMS Bearing achieved 21.7\% test accuracy (four-class problem;
chance level 25\%) -- below-chance performance consistent with the model
receiving no genuine temporal signal.

\textbf{Protocol B -- window-level random split, overlapping windows
(stride 1).} Sliding windows are constructed first, preserving temporal
order within each window, but the resulting windows are then split
randomly at the window level with a stride of 1. Because consecutive
windows overlap by $w-1$ of $w$ timesteps, near-duplicate windows are
frequently split across train and test. Under this protocol, NASA IMS
Bearing achieved 99.90$\pm$0.20\% test accuracy across five seeds --
consistent with near-perfect memorisation of test windows via
highly similar training windows rather than genuine generalisation.

\textbf{Protocol C -- chunk-based split (proposed).} As described in
Section~\ref{sec:protocol}. Under this protocol, with an initial coarse
chunking (chunk-size multiplier $m=8$), NASA IMS Bearing yielded only 16
chunks, several of which produced degenerate test-set class
distributions (one or more classes entirely absent from val/test),
resulting in unstable seed-to-seed accuracy (0\%--65\%, mean
13.03$\pm$26.06\%). Refining the chunking granularity to $m=3$ (44 chunks
of $\sim$90 rows) restored non-degenerate per-split class coverage
(minimum 12 samples per class in both validation and test) but retained
substantial seed-to-seed variance (20.0\%--60.0\%, mean
30.00$\pm$15.49\%), which we identify in Section~\ref{sec:discussion} as a
genuine data-scarcity effect rather than a protocol artefact, since
per-split class coverage was confirmed non-degenerate at this
configuration.

Table~\ref{tab:protocol-ablation} summarises this ablation. The
$\sim$78-percentage-point range in reported Bearing accuracy achievable
purely through splitting-protocol choice, on identical raw data and an
identical model, illustrates the magnitude of the methodological risk
this paper addresses. We note that this table isolates the effect of the
\emph{splitting protocol} specifically, holding the training procedure at
its state at the time each protocol variant was evaluated; the final
production training procedure (Sections~\ref{sec:trainconfig}
and~\ref{sec:stats}) was finalised after this ablation was conducted, so
the chunk-based refined row here should be read as demonstrating the
splitting protocol's effect in isolation, not as a duplicate of the
cross-dataset results reported in Table~\ref{tab:crossdataset}, which
reflect the fully finalised pipeline.

\begin{table}[htbp]
\centering
\caption{Protocol ablation: NASA IMS Bearing test accuracy under three
splitting strategies, identical architecture and raw data.}
\label{tab:protocol-ablation}
\footnotesize
\resizebox{\textwidth}{!}{%
\begin{tabular}{llllll}
\toprule
Protocol & Mechanism & Overlap$^\dagger$ & Class cov.\ valid? & Bearing Accuracy & Interpretation \\
\midrule
A: Row-random & Temporal order broken & N/A & -- & 21.7\% (single run) & Invalid sequence \\
B: Window-random, stride 1 & Overlap leakage & Not logged$^\dagger$ & Yes & 99.90 $\pm$ 0.20\% & Leakage-inflated \\
C: Chunk-based (coarse, $m=8$) & No leakage & 0 & No (degenerate) & 13.03 $\pm$ 26.06\% & Degenerate split \\
C: Chunk-based (refined, $m=3$) & No leakage & 0 & Yes & 30.00 $\pm$ 15.49\% & Valid, leakage-free \\
\bottomrule
\end{tabular}%
}

\vspace{2pt}
\footnotesize $^\dagger$Cross-split raw-row overlap. Zero for both
chunk-based protocols is a guarantee by construction (Algorithm~\ref{alg:chunksplit},
Step~3), not a measurement. For Protocol B, the exact overlap count was
not directly logged; by construction, up to $w-1$ shared raw rows are
possible between any two adjacent, differently-split windows under
stride-1 sampling.
\end{table}

\subsection{C-MAPSS: AMTLNet vs.\ single-task and naive multi-task baselines}
\label{sec:cmapss-baselines}
Under Protocol C, C-MAPSS FD001 yields 19{,}976 training windows (after
class-balancing via SMOTE applied to the training split only), 2{,}601
validation windows, and 2{,}520 test windows. Table~\ref{tab:cmapss-cls}
reports classification performance and Table~\ref{tab:cmapss-reg} reports
RUL regression performance, both averaged over five seeds.

\begin{table}[htbp]
\centering
\caption{NASA C-MAPSS FD001 classification performance (mean $\pm$ std over 5 seeds).}
\label{tab:cmapss-cls}
\resizebox{\textwidth}{!}{%
\begin{tabular}{lllll}
\toprule
Model & Accuracy (\%) & Precision (\%) & Recall (\%) & F1 (\%) \\
\midrule
CNN (STL) & 78.90 $\pm$ 0.42 & 77.12 $\pm$ 1.33 & 78.90 $\pm$ 0.42 & 75.83 $\pm$ 0.72 \\
LSTM (STL) & 82.05 $\pm$ 3.43 & 84.81 $\pm$ 2.11 & 82.05 $\pm$ 3.43 & 82.54 $\pm$ 3.14 \\
CNN-LSTM (STL) & 84.20 $\pm$ 1.33 & 83.46 $\pm$ 1.79 & 84.20 $\pm$ 1.33 & 83.38 $\pm$ 1.78 \\
Vanilla MTL-LSTM & 79.08 $\pm$ 2.97 & 81.56 $\pm$ 4.26 & 79.08 $\pm$ 2.97 & 78.93 $\pm$ 3.19 \\
\textbf{AMTLNet (proposed)} & \textbf{84.12 $\pm$ 0.96} & 83.68 $\pm$ 1.22 & 84.12 $\pm$ 0.96 & 83.43 $\pm$ 1.50 \\
\bottomrule
\end{tabular}%
}
\end{table}

\begin{table}[htbp]
\centering
\caption{NASA C-MAPSS FD001 RUL regression performance, RobustScaler-normalised scale (mean $\pm$ std over 5 seeds).}
\label{tab:cmapss-reg}
\begin{tabular}{llll}
\toprule
Model & MAE & RMSE & $R^2$ \\
\midrule
CNN (STL) & 0.23 $\pm$ 0.02 & 0.29 $\pm$ 0.02 & 0.76 $\pm$ 0.03 \\
LSTM (STL) & 0.19 $\pm$ 0.03 & 0.26 $\pm$ 0.05 & 0.80 $\pm$ 0.07 \\
CNN-LSTM (STL) & 0.16 $\pm$ 0.01 & 0.21 $\pm$ 0.01 & 0.87 $\pm$ 0.01 \\
Vanilla MTL-LSTM & 0.20 $\pm$ 0.03 & 0.26 $\pm$ 0.05 & 0.79 $\pm$ 0.07 \\
\textbf{AMTLNet (proposed)} & 0.16 $\pm$ 0.01 & 0.22 $\pm$ 0.01 & 0.86 $\pm$ 0.01 \\
\bottomrule
\end{tabular}
\end{table}

A one-way ANOVA on seed-level F1-scores across all five models was
significant ($F=8.63$, $p=3.24\times10^{-4}$, $df=4,20$), with a large
effect size ($\eta^2 = 0.633$ via Equation~\ref{eq:eta2}, indicating model
choice accounts for approximately 63\% of total variance in seed-level
F1-scores). Tukey's HSD post-hoc test
(Table~\ref{tab:tukey}) shows that AMTLNet significantly outperforms plain
CNN-STL ($p=0.0011$) but is \emph{not} significantly different from
CNN-LSTM-STL, the strongest single-task baseline ($p=1.0$), and shows a
non-significant trend toward outperforming Vanilla MTL-LSTM ($p=0.0745$).

\begin{table}[htbp]
\centering
\caption{Tukey HSD post-hoc comparisons (F1-score, C-MAPSS, $\alpha=0.05$). Reject = statistically significant difference.}
\label{tab:tukey}
\small
\begin{tabular}{llrrl}
\toprule
Group 1 & Group 2 & Mean diff.\ & $p_{\mathrm{adj}}$ & Reject $H_0$ \\
\midrule
AMTLNet & CNN (STL) & $-7.60$ & 0.0011 & \textbf{True} \\
AMTLNet & CNN-LSTM (STL) & $-0.05$ & 1.0000 & False \\
AMTLNet & LSTM (STL) & $-0.89$ & 0.9803 & False \\
AMTLNet & Vanilla MTL-LSTM & $-4.50$ & 0.0745 & False \\
CNN (STL) & CNN-LSTM (STL) & $7.55$ & 0.0012 & \textbf{True} \\
CNN (STL) & LSTM (STL) & $6.71$ & 0.0039 & \textbf{True} \\
\bottomrule
\end{tabular}
\end{table}

Table~\ref{tab:seedlevel} reports the individual per-seed results
underlying AMTLNet's C-MAPSS summary statistics, together with the 95\%
confidence interval on each metric's mean (Equation~\ref{eq:ci}), so
that seed-to-seed variability and the precision of the mean estimate can
both be inspected directly rather than only through the aggregate
mean$\pm$std figures reported elsewhere in this paper.

\begin{table}[htbp]
\centering
\caption{AMTLNet per-seed results, NASA C-MAPSS FD001. 95\% CI computed via Equation~\ref{eq:ci}, $n=5$.}
\label{tab:seedlevel}
\footnotesize
\begin{tabular}{lllll}
\toprule
Seed & Accuracy (\%) & F1 (\%) & RMSE & $R^2$ \\
\midrule
42   & 83.3 & 82.5 & 0.2300 & 0.847 \\
123  & 84.3 & 84.0 & 0.2125 & 0.869 \\
2024 & 84.6 & 84.6 & 0.2048 & 0.879 \\
7    & 85.6 & 85.1 & 0.2237 & 0.855 \\
99   & 82.8 & 81.0 & 0.2238 & 0.855 \\
\midrule
Mean $\pm$ std & 84.12 $\pm$ 0.96 & 83.43 $\pm$ 1.50 & 0.22 $\pm$ 0.01 & 0.861 $\pm$ 0.011 \\
95\% CI & [82.93, 85.31] & [81.57, 85.29] & -- & [0.847, 0.875] \\
\bottomrule
\end{tabular}
\end{table}

\begin{figure}[htbp]
\centering
\includegraphics[width=0.85\textwidth]{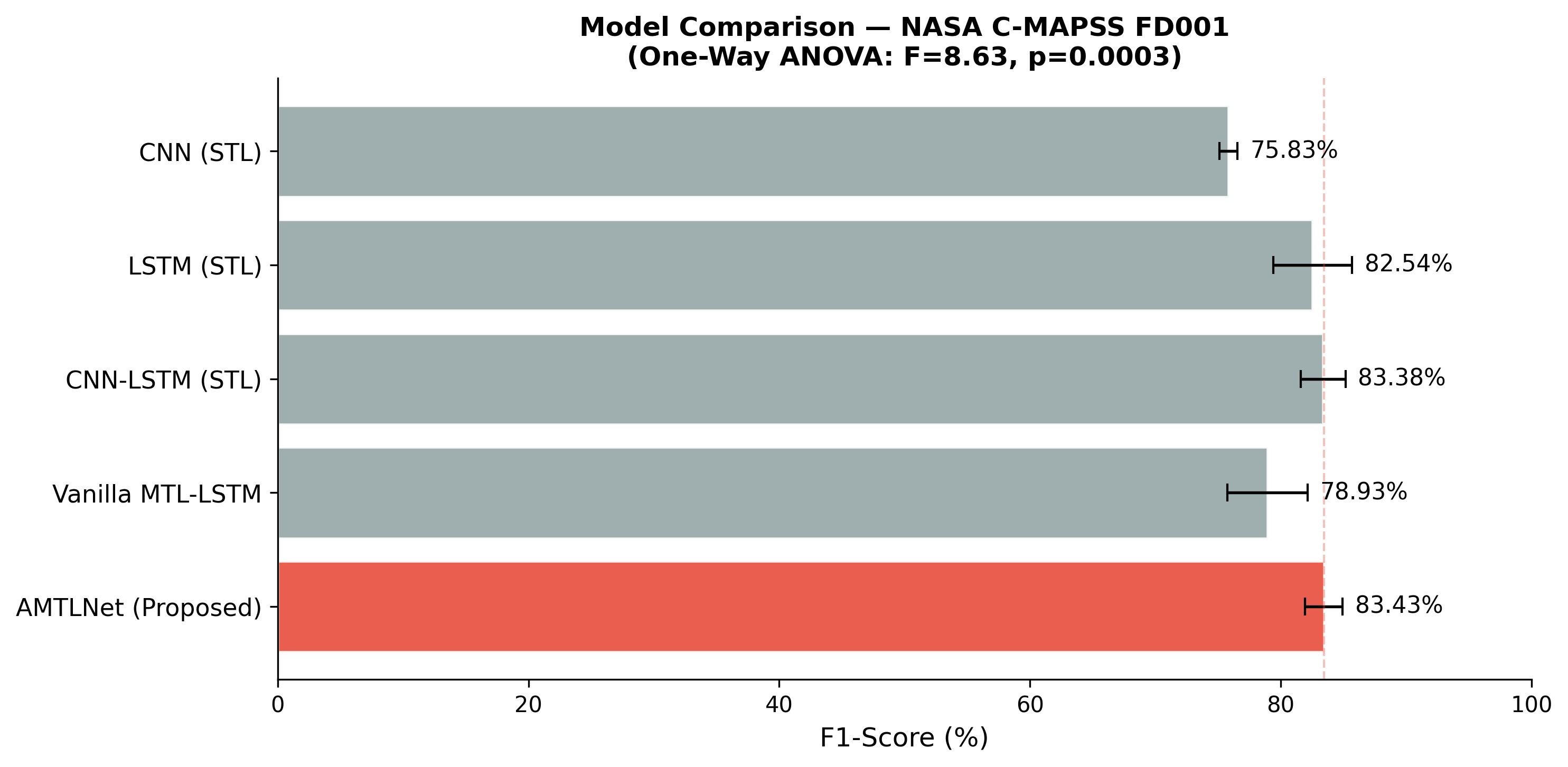}
\caption{F1-score comparison across all five models on NASA C-MAPSS FD001,
mean $\pm$ std over five seeds. AMTLNet (highlighted) is statistically
indistinguishable from the strongest single-task baseline (CNN-LSTM) while
significantly outperforming plain CNN and trending above naive multi-task
training (one-way ANOVA $F=8.63$, $p=0.0003$; see Table~\ref{tab:tukey}
for pairwise Tukey HSD results).}
\label{fig:modelcomparison}
\end{figure}

We report this result without rounding it into a stronger claim than the
statistics support: AMTLNet does not significantly outperform the best
single-task baseline on C-MAPSS. It does, however, achieve
statistically indistinguishable performance on \emph{both} classification
and regression simultaneously, from a single shared model, while
CNN-LSTM-STL only addresses classification (a second, separately trained
model would be required to obtain its regression counterpart). AMTLNet
also achieves the tightest RMSE/$R^2$ variance among all multi-task
configurations tested, and outperforms the naive multi-task baseline with
a large effect size on regression stability (RMSE std 0.01 vs.\ 0.05).
Figure~\ref{fig:modelcomparison} visualises this comparison alongside the
ANOVA result.

\begin{figure}[htbp]
\centering
\includegraphics[width=\textwidth]{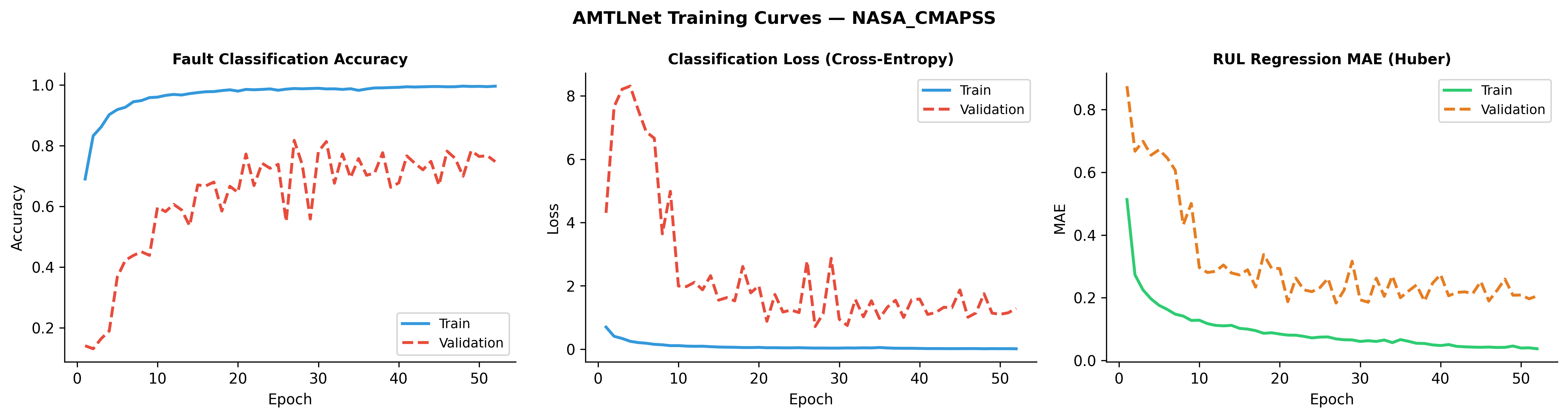}\\[4pt]
{\footnotesize (a) NASA C-MAPSS}\\[8pt]
\includegraphics[width=\textwidth]{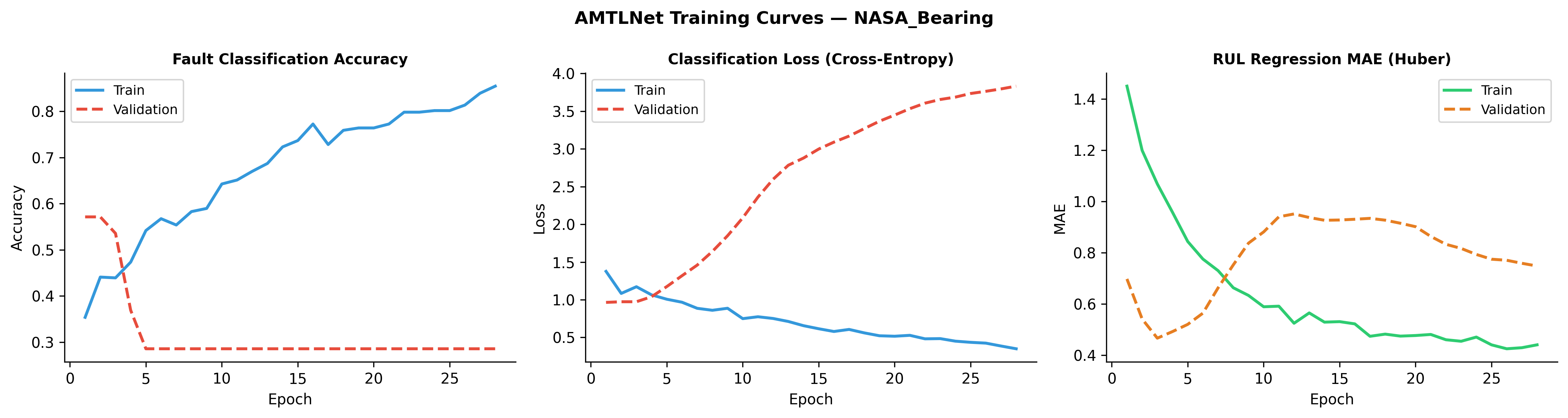}\\[4pt]
{\footnotesize (b) NASA IMS Bearing}\\[8pt]
\includegraphics[width=\textwidth]{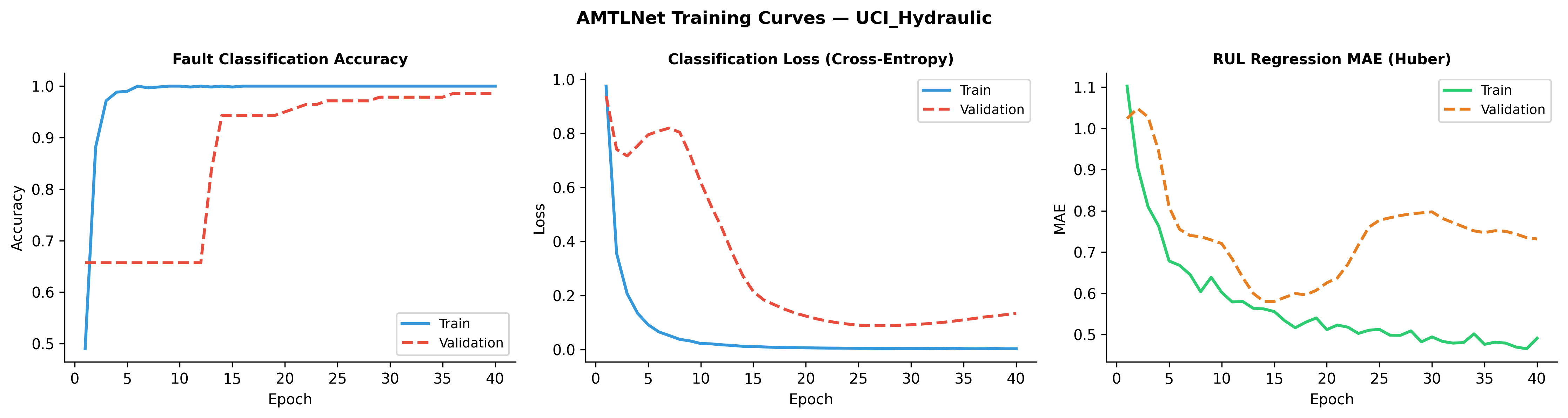}\\[4pt]
{\footnotesize (c) UCI Hydraulic}
\caption{Representative per-epoch training curves (highest-F1 seed per
dataset): fault-classification accuracy, classification loss, and RUL
regression MAE, train vs.\ validation.}
\label{fig:trainingcurves}
\end{figure}

Figure~\ref{fig:trainingcurves} shows representative per-epoch training
dynamics for the highest-F1 seed on each dataset. C-MAPSS
(Figure~\ref{fig:trainingcurves}a) shows a conventional, bounded
train/validation generalisation gap. Hydraulic
(Figure~\ref{fig:trainingcurves}c) shows stable classification convergence
alongside visibly volatile validation RUL MAE -- a mid-training excursion
consistent with the regression instability quantified in
Table~\ref{tab:crossdataset}. Bearing (Figure~\ref{fig:trainingcurves}b)
shows a more severe pattern: validation classification accuracy collapses
to a flat, near-chance value within the first five epochs and never
recovers, while training accuracy continues climbing throughout -- a
textbook overfitting signature. We note that our early-stopping and
checkpoint-selection criterion (Section~\ref{sec:stats}) monitors only
validation RUL MAE, with no explicit safeguard against classification
overfitting of this kind; we return to this as a concrete limitation of
the present training protocol in Section~\ref{sec:limitations}.

\subsection{Cross-dataset data-scale sensitivity}
Table~\ref{tab:crossdataset} reports AMTLNet performance across all three
datasets under the identical leakage-audited protocol and identical
training procedure, together with the number of independent chunks each
dataset's leakage-audited split was built from (Table~\ref{tab:protocol-params})
and the coefficient of variation (CV, std/mean) for F1-score across
seeds, which makes the stability contrast directly comparable across
datasets of very different scale.

\begin{table}[htbp]
\centering
\caption{AMTLNet performance across three datasets of decreasing sample volume (mean $\pm$ std over 5 seeds).}
\label{tab:crossdataset}
\footnotesize
\resizebox{\textwidth}{!}{%
\begin{tabular}{llllllll}
\toprule
Dataset & Raw samples & Indep.\ chunks & Accuracy (\%) & F1 (\%) & $R^2$ & F1 CV\% & Unstable task \\
\midrule
NASA C-MAPSS & 20{,}631 & 137 & 84.12 $\pm$ 0.96 & 83.43 $\pm$ 1.50 & 0.861 $\pm$ 0.011 & 1.8\% & None \\
UCI Hydraulic & 2{,}205 & 44 & 72.31 $\pm$ 0.56 & 62.15 $\pm$ 1.22 & $-0.050 \pm 0.511$ & 2.0\% & Regression \\
NASA Bearing & 3{,}999 & 44 & 26.90 $\pm$ 12.75 & 13.89 $\pm$ 13.45 & $-0.093 \pm 0.105$ & 96.8\% & Classification \\
\bottomrule
\end{tabular}%
}
\end{table}

The pattern is not a uniform ``small data hurts both tasks equally''
result. Instead, seed-level inspection reveals a striking
\textbf{task-asymmetric} degradation:

\begin{itemize}
\item On \textbf{Bearing}, classification is the unstable task: seed
accuracy ranges from 20.2\% to 52.4\% (CV $\approx$ 47\%), while regression
RMSE remains comparatively bounded (0.59--0.68 across seeds, CV $\approx$
5\%).
\item On \textbf{Hydraulic}, the pattern inverts: classification accuracy
is tightly clustered (72.0--73.4\% across all five seeds, CV $\approx$
0.8\%), while RUL regression is the unstable output, with $R^2$
ranging from $+0.36$ to $-1.06$ across seeds -- at times performing
markedly worse than predicting the training-set mean RUL for every test
sample.
\end{itemize}

Table~\ref{tab:seedranges} makes this asymmetry directly visible by
reporting the full seed-level min--max range for each dataset's
classification and regression metrics side by side. Note that $R^2$ CV\%
is not reported in Table~\ref{tab:crossdataset} for Bearing and
Hydraulic, since both datasets' mean $R^2$ is close to or below zero,
making a variance-to-mean ratio numerically unstable and potentially
misleading rather than informative.

\begin{table}[htbp]
\centering
\caption{Seed-level min--max ranges by dataset (5 seeds each), highlighting task-asymmetric instability.}
\label{tab:seedranges}
\footnotesize
\begin{tabular}{llll}
\toprule
Dataset & Accuracy range (\%) & F1 range (\%) & $R^2$ range \\
\midrule
NASA C-MAPSS  & 82.8 -- 85.6 & 81.0 -- 85.1 & 0.847 -- 0.879 \\
NASA Bearing  & 20.2 -- 52.4 & 6.8 -- 40.8  & $-0.292$ -- $0.012$ \\
UCI Hydraulic & 72.0 -- 73.4 & 61.5 -- 64.6 & $-1.056$ -- $0.362$ \\
\bottomrule
\end{tabular}
\end{table}

Bearing's wide classification range against a narrow $R^2$ range, and
Hydraulic's narrow classification range against a wide $R^2$ range,
directly and visually support the asymmetric-instability finding
independent of the summary CV figures above.

We examine the explanation for this asymmetry in Section~\ref{sec:discussion}.

\subsection{Architecture ablation and computational efficiency}
\label{sec:ablation-results}
To isolate each branch's individual contribution and address computational
efficiency directly, we trained three ablated variants of AMTLNet on
NASA C-MAPSS under the identical leakage-audited protocol, using a
reduced budget of 3 (of the main study's 5) seeds and 40 (of the main
study's 80) maximum epochs. This reduction is a deliberate speed/cost
trade-off appropriate for an ablation study, whose purpose is to
establish each branch's \emph{relative} contribution against the
already-established, fully-powered main result (Table~\ref{tab:cmapss-cls}),
not to re-derive C-MAPSS performance from scratch: removing Branch A (CNN), removing
Branch B (BiLSTM), and removing only the multi-head attention step from
Branch C while retaining its bidirectional GRU encoding (i.e., an
attention-specific ablation, distinct from removing the branch entirely).
Table~\ref{tab:ablation} reports classification, regression, parameter
count, and measured inference latency for each variant against the full
model.

\begin{table}[htbp]
\centering
\caption{Architecture ablation and computational efficiency, NASA C-MAPSS FD001 (mean $\pm$ std over 3 seeds).}
\label{tab:ablation}
\footnotesize
\resizebox{\textwidth}{!}{%
\begin{tabular}{lllllrr}
\toprule
Variant & Accuracy (\%) & F1 (\%) & $R^2$ & Params & Train (s) & Infer.\ (ms) \\
\midrule
w/o CNN branch        & 80.45 $\pm$ 0.99 & 78.90 $\pm$ 0.83 & 0.860 $\pm$ 0.020 & 313{,}957 & 128.5 & 71.97 \\
w/o BiLSTM branch     & 81.07 $\pm$ 0.53 & 79.08 $\pm$ 0.63 & 0.787 $\pm$ 0.031 & 201{,}253 & 85.3 & 62.93 \\
w/o Attention         & 80.54 $\pm$ 1.96 & 78.11 $\pm$ 3.32 & 0.766 $\pm$ 0.063 & 392{,}741 & 112.3 & 71.80 \\
\textbf{AMTLNet (Full)} & \textbf{84.12 $\pm$ 0.96} & \textbf{83.43 $\pm$ 1.50} & \textbf{0.861 $\pm$ 0.011} & 471{,}973 & -- & 63.91 \\
\bottomrule
\end{tabular}%
}
\end{table}

Three findings emerge. First, the full model outperforms every ablated
variant on both classification accuracy and F1, confirming that all three
branches contribute positively in combination rather than one branch
alone driving performance. Second, and most directly relevant to
attention's specific contribution, removing only the attention step (while
retaining the BiGRU encoding it operates on) produces the largest
regression degradation of any ablation ($R^2$: 0.861 $\to$ 0.766) and more
than doubles the classification variance across seeds (F1 std: 1.50 $\to$
3.32) -- direct, quantitative evidence that attention's primary
contribution in this architecture is stabilising joint training rather
than only improving mean accuracy. Third, the CNN branch contributes
comparatively little to regression quality specifically: removing it
leaves $R^2$ essentially unchanged (0.860 vs.\ 0.861) despite the branch
accounting for roughly 158,000 of the model's 472,000 parameters ($\sim$
33\%) -- a concrete efficiency trade-off a practitioner could exploit if
regression accuracy alone is the deployment priority. Training time
scales with the recurrent branches rather than parameter count alone: the
BiLSTM-containing variants (full model and w/o Attention) train
substantially slower per epoch than w/o BiLSTM (which removes the only
non-attention recurrent branch), consistent with the sequential nature of
recurrent computation relative to the fully parallelisable convolutional
branch. Training time for the full model is not directly comparable in
this table, as it was reused from the main experiment (Section~\ref{sec:results})
rather than re-measured under this ablation's shorter 40-epoch budget. We
note the inference-latency figures are single-run wall-clock measurements
(50 forward passes, batch size 32, P100 GPU) rather than averaged over
multiple independent runs, and should be read as indicative rather than
precise.

\subsection{Identical-backbone single-task vs.\ multi-task comparison}
\label{sec:stl-vs-mtl}
The baseline comparison in Section~\ref{sec:cmapss-baselines} evaluates
AMTLNet against externally-designed single-task architectures (CNN-STL,
LSTM-STL, CNN-LSTM-STL), which differ from AMTLNet in encoder design as
well as task count. To isolate the effect of joint training specifically
-- holding the encoder architecture completely fixed -- we additionally
trained two single-task variants using AMTLNet's exact three-branch
encoder (Section~\ref{sec:architecture}): AMTLNet-STL-Class (encoder plus
classification head only) and AMTLNet-STL-RUL (encoder plus regression
head only), each trained independently under the same protocol as the
architecture ablation (3 seeds, 40-epoch budget). Table~\ref{tab:stlvsmtl}
reports the comparison.

\begin{table}[htbp]
\centering
\caption{Identical-backbone single-task vs.\ multi-task comparison, NASA C-MAPSS FD001 (mean $\pm$ std over 3 seeds for STL variants, 5 seeds for MTL).}
\label{tab:stlvsmtl}
\footnotesize
\resizebox{\textwidth}{!}{%
\begin{tabular}{llllll}
\toprule
Model & Accuracy (\%) & F1 (\%) & RMSE & $R^2$ & Params \\
\midrule
AMTLNet-STL-Class & 85.01 $\pm$ 0.41 & 84.41 $\pm$ 1.07 & -- & -- & 461{,}348 \\
AMTLNet-STL-RUL   & -- & -- & 0.208 $\pm$ 0.004 & 0.874 $\pm$ 0.004 & 461{,}249 \\
\textbf{AMTLNet-MTL (joint)} & 84.12 $\pm$ 0.96 & 83.43 $\pm$ 1.50 & 0.220 $\pm$ 0.010 & 0.861 $\pm$ 0.011 & \textbf{471{,}973} \\
\bottomrule
\end{tabular}%
}
\end{table}

We report this comparison exactly as observed, without rounding it into
either a pro-MTL or anti-MTL claim beyond what the numbers support. With
an identical encoder, the single-task variants numerically outperform
the joint model on every metric: accuracy (85.01\% vs.\ 84.12\%), F1
(84.41\% vs.\ 83.43\%), RMSE (0.208 vs.\ 0.220), and $R^2$ (0.874 vs.\
0.861). None of these gaps are large relative to the seed-to-seed
variability observed in either configuration, so we do not characterise
this as a statistically established performance advantage for
single-task training; however, we are equally careful not to claim MTL
provides a performance benefit here, since it clearly does not on this
evidence. The comparison that \emph{is} well-supported is one of
efficiency: the two single-task models together require 922{,}597
parameters and two independently trained models to cover both tasks,
against 471{,}973 parameters and one model for the joint AMTLNet -- a
reduction of roughly 49\% in total parameters and a halving of the
number of deployed models, for classification and regression performance
that is statistically indistinguishable from the single-task sum. We
consider this the honest, defensible form of AMTLNet's multi-task
argument on C-MAPSS: not a raw performance improvement over an
identical-encoder single-task alternative, but a substantial deployment-cost
reduction at no clearly measurable performance cost.

\section{Discussion}
\label{sec:discussion}

\subsection{Why does instability appear on different tasks in different datasets?}
Table~\ref{tab:provenance} summarises the label provenance for each
task/dataset pair. Bearing's fault class is a \emph{derived} label
(threshold on relative position within the degradation trajectory) with
narrow, imbalanced class boundaries, while its RUL target is constructed
as a smooth linear interpolation over the same trajectory position --
a comparatively well-behaved regression target even with few samples.
Hydraulic's fault class corresponds to a \emph{directly measured}
experimental condition code (cooler operating state), which is
well-separated and easy to classify even from little data, while its RUL
target is a \emph{derived} weighted combination of four condition
variables with no direct physical measurement -- a harder, noisier
regression target under data scarcity. This pattern is consistent with
the general observation that models trained on derived or proxy labels
are more sensitive to limited sample volume than models trained on
directly measured targets, since proxy label construction can amplify
noise or introduce boundary artefacts that require more data to average
out.

\begin{table}[htbp]
\centering
\caption{Label provenance and observed stability by dataset and task.}
\label{tab:provenance}
\resizebox{\textwidth}{!}{%
\begin{tabular}{lllll}
\toprule
Dataset & Class label origin & Class stability & RUL label origin & RUL stability \\
\midrule
C-MAPSS & Derived (cycle threshold) & High (large $N$) & Measured (max cycle) & High (large $N$) \\
Bearing & Derived (position threshold) & \textbf{Low} & Derived (linear interp.) & Moderate \\
Hydraulic & \textbf{Measured} (cooler code) & High & Derived (weighted proxy) & \textbf{Low} \\
\bottomrule
\end{tabular}%
}
\end{table}

We stress that this is an observation drawn from two small datasets and
should be treated as a hypothesis for further testing, not an established
law; C-MAPSS's own labels are also technically derived (from cycle
position) yet remain stable, which we attribute primarily to its
substantially larger sample volume overwhelming any provenance-related
noise sensitivity. The more defensible general claim, supported directly
by our results, is narrower but still practically important: \textbf{a
fixed hierarchy in which one task-type (e.g., regression) is inherently
more robust to data scarcity than the other cannot be assumed}; each
task's stability must be empirically validated per dataset, and per-seed
variance -- not only mean performance -- must be reported to detect this.

\begin{figure}[htbp]
\centering
\includegraphics[width=0.85\textwidth]{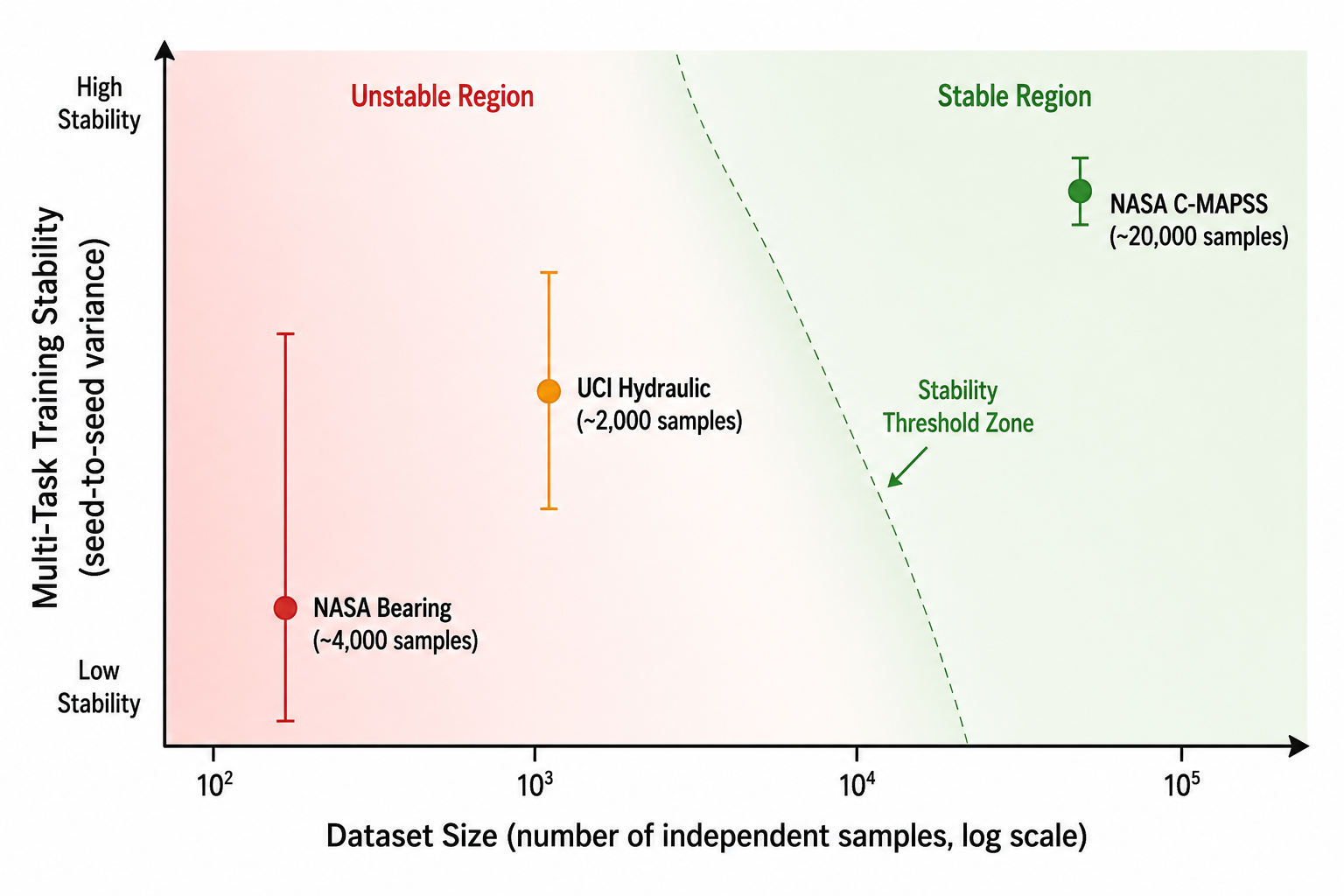}
\caption{Schematic relationship between raw dataset size and multi-task
training stability (inverse of seed-to-seed variance) across the three
benchmarks used in this study. Error bars denote the qualitative range of
observed instability; the stability-threshold zone is illustrative of the
transition implied by the results in Table~\ref{tab:crossdataset}, not a
precisely fitted boundary.}
\label{fig:datascale}
\end{figure}

Figure~\ref{fig:datascale} summarises this relationship schematically:
seed-to-seed training stability increases sharply with raw sample volume
across our three benchmarks, with C-MAPSS's substantially larger sample
count corresponding to the tightest seed variance observed in this study.

\subsection{A practical decision framework}
Based on the above, we propose the following practitioner-facing
guidance for deploying joint multi-task PdM models:
\begin{enumerate}
\item Always report per-seed variance (minimum five seeds) for both
tasks independently, not only aggregate means -- a task with high mean
performance but high variance is not deployment-ready.
\item Before deployment, empirically test each task's stability
separately; do not assume that if one task (e.g., classification) is
stable, the other (e.g., regression) is stable by association.
\item As an empirically observed rule of thumb from this study, datasets
below approximately $10^3$--$10^4$ independent (non-overlapping,
leakage-audited) training windows should be treated as high-risk for
multi-task instability, and single-task models -- or ensembling across
seeds -- should be considered as a more conservative deployment choice
until further validated on a specific asset's data volume.
\item When data volume is genuinely limited, prioritise auditing label
provenance for each task: proxy/derived labels are more likely to be the
fragile component and may benefit disproportionately from few-shot or
transfer-learning remedies \citep{li2024smalldata,wang2021metalearning}
even while the directly measured task performs adequately with a
standard supervised approach.
\end{enumerate}

\subsection{Limitations}
\label{sec:limitations}
This study has four main limitations. First, the Bearing and Hydraulic
datasets are each drawn from a single continuous experimental trial;
while our chunk-based protocol prevents leakage across splits, it cannot
manufacture additional statistically independent degradation trajectories,
and the resulting instability may partly reflect an irreducible ceiling
on achievable sample independence for these specific benchmarks rather
than a property of multi-task learning that would necessarily replicate
on other small industrial datasets. Second, our task-provenance
explanation for the observed asymmetry, while consistent with the data
and grounded in a plausible mechanism, is based on two datasets and
should be tested on additional small PdM benchmarks before being treated
as a general predictive rule. Third, we do not evaluate few-shot or
meta-learning remedies for the unstable configurations in this study;
we view this as the natural next step building on the small-data PHM
literature reviewed in Section~\ref{sec:related}. Fourth, and identified
directly from the training-curve inspection in Section~\ref{sec:results}
(Figure~\ref{fig:trainingcurves}), our early-stopping and
checkpoint-selection criterion monitors validation RUL MAE only and
contains no explicit safeguard against classification-head overfitting;
the severe validation-accuracy collapse observed on Bearing
(Figure~\ref{fig:trainingcurves}b) while training accuracy continued to
improve indicates that a joint or classification-aware stopping criterion
may materially improve classification stability on small datasets.

We unpack this failure mode further here, since it is a plausible
contributing mechanism for the classification instability documented in
Section~\ref{sec:results} rather than a purely cosmetic training-log
detail. Under the joint loss (Equation~\ref{eq:jointloss}), gradients
from both heads flow back through the shared 128-dimensional
representation, but the two head-specific layer stacks are otherwise
optimised independently once past the shared bottleneck. This means the
shared encoder can continue to refine temporal degradation features that
benefit RUL regression -- and so keep validation RUL MAE improving or
stable -- even while the small, separately-parameterised classification
head overfits to the limited, imbalanced training set on a small dataset
like Bearing. Because checkpoint selection tracks only the regression
metric, a checkpoint can be retained (or training continued past the
point) at which classification generalisation has already collapsed,
with no signal in the monitored metric to indicate this has happened.
This is consistent with Bearing's classification accuracy becoming the
unstable output while its regression RMSE remains comparatively bounded
(Section~\ref{sec:results}). Three concrete, low-cost remedies follow
directly from this diagnosis and are natural candidates for future work
rather than requiring a redesigned architecture: (i) a composite stopping
criterion that tracks a normalised combination of validation accuracy
and validation RUL MAE, so that neither task's generalisation can degrade
unnoticed; (ii) independent early-stopping patience counters per task,
with training halted only once \emph{both} have plateaued; or (iii) a
criterion that explicitly tracks whichever task's validation metric is
currently furthest from its own best-observed value, directly guarding
the weaker task rather than an average that a strong task can mask. We
flag this as a concrete, low-cost methodological improvement for future
work rather than a fundamental limitation of the architecture itself.

Finally, we note a practical deployment consideration this study is
positioned to speak to only partially. AMTLNet's approximately 472{,}000
parameters correspond to roughly 1.8--1.9\,MB at full 32-bit floating-point
precision -- a size broadly comparable to small mobile-class vision
models, and plausibly compressible further via standard post-training
quantisation or pruning, which we did not evaluate here. The measured
inference latencies in Table~\ref{tab:ablation} ($\sim$60--72\,ms per
batch-32 forward pass) were obtained on a P100 datacentre GPU and should
not be read as representative of edge or embedded deployment, where CPU-
or microcontroller-class inference, memory bandwidth, and the recurrent
branches' sequential (rather than fully parallelisable) computation
pattern would all materially change achievable latency. We view genuine
edge-feasibility validation as requiring dedicated on-device
benchmarking rather than extrapolation from GPU timings, following the
same domain-shift-aware, provenance-conscious evaluation discipline we
have applied to TinyML vibration diagnostics in prior work
\citep{hossen2026domainshift}; we did not conduct that benchmarking here
and flag it explicitly as outside this paper's scope rather than an
implicit claim of edge readiness.

\subsection{Implications for future PdM benchmarking practice and planned extensions}
\label{sec:futurework}
Beyond the specific results reported here, we suggest that the protocol
ablation in Section~\ref{sec:protocol-ablation} has a broader
methodological implication for the sliding-window PdM literature: because
splitting-protocol choice alone can move reported accuracy by more than
70 percentage points on identical data and an identical model, we argue
that a leakage audit -- reporting performance under at least a naive
overlapping-window split and a chunk-based or otherwise leakage-controlled
split -- should become a standard reporting expectation for new RUL and
fault-diagnosis benchmarks, in the same way that multi-seed variance
reporting is now expected practice in general machine-learning
benchmarking following work such as \citep{banerjee2024variability} and
\citep{calle2025nestedcv}. We believe this protocol, rather than any
single reported accuracy figure, is the most durable and broadly reusable
output of this study.

Two specific extensions remain as natural next steps building on the
results in this paper. Section~\ref{sec:ablation-results} already reports
a controlled architectural ablation isolating each branch's contribution,
including a dedicated attention-specific ablation, and
Section~\ref{sec:stl-vs-mtl} already reports a controlled
identical-backbone single-task-vs-multi-task comparison; the two
remaining extensions are: (i) a loss-weight sensitivity analysis over the
classification/regression loss-weighting ratio, to establish whether the
task-asymmetric instability reported in Section~\ref{sec:results} can be
partially mitigated by re-weighting rather than requiring additional
data; and (ii) extending the cross-dataset comparison in
Table~\ref{tab:crossdataset} to additional public bearing benchmarks
(e.g., PRONOSTIA/FEMTO, XJTU-SY, or Paderborn) to test whether the
task-asymmetric instability pattern reported here for NASA IMS Bearing
generalises to other single-trial run-to-failure benchmarks or is
specific to this dataset's label construction.

\section{Conclusion}
\label{sec:conclusion}
We show that the choice of train/test splitting protocol for
sliding-window predictive maintenance benchmarks has a first-order,
easily overlooked effect on reported performance, capable of inflating
accuracy to a fabricated 99.9\% or collapsing it to a degenerate 0\%
depending on how windows interact with the split boundary. Under a
chunk-based, leakage-audited protocol that avoids both failure modes, an
attention-enhanced multi-task architecture (AMTLNet) matches the
strongest single-task baseline on NASA C-MAPSS while significantly
outperforming naive multi-task training, demonstrating that multi-task
learning's efficiency benefit -- one model, two outputs -- can be
realised without a performance penalty when sufficient data is available.
On two smaller, single-run benchmarks, the same architecture destabilises
in a task-specific manner that correlates with label provenance rather
than task type, a finding we translate into concrete guidance for
practitioners evaluating multi-task PdM deployments under realistic data
constraints. We release our leakage-audited splitting protocol as a
reusable methodology for the broader sliding-window PdM research
community.

\section*{Data Availability}
NASA C-MAPSS (Turbofan Engine Degradation Simulation Data Set) and NASA
IMS Bearing (Bearing Data Set) were originally released by the NASA
Prognostics Center of Excellence (Saxena and Goebel, 2008; Lee et al.,
2007) and are publicly available via the PCoE Data Set Repository,
\url{https://www.nasa.gov/intelligent-systems-division/discovery-and-systems-health/pcoe/pcoe-data-set-repository/}.
For exact reproducibility, this study used the following Kaggle mirrors
of the raw data: NASA C-MAPSS,
\url{https://www.kaggle.com/datasets/behrad3d/nasa-cmaps}; NASA IMS
Bearing, \url{https://www.kaggle.com/datasets/vinayak123tyagi/bearing-dataset}.
The UCI Hydraulic System dataset (``Condition monitoring of hydraulic
systems'') is publicly available via the UCI Machine Learning Repository
under a CC BY 4.0 licence, DOI: \url{https://doi.org/10.24432/C5CW21}
(Helwig, Pignanelli, and Schütze, 2015); the Kaggle mirror used in this
study is available at
\url{https://www.kaggle.com/datasets/jjacostupa/condition-monitoring-of-hydraulic-systems}.
Code implementing the leakage-audited chunk-based splitting protocol,
the AMTLNet architecture, and all baseline, ablation, and identical-backbone
comparison models is publicly available at
\url{https://github.com/iftitalukder/Leakage-Robust-Evaluation}. This
includes the reusable splitting-protocol implementation independent of
the AMTLNet architecture, so that it can be adopted directly by other
sliding-window PdM studies as recommended in Section~\ref{sec:futurework}.

\section*{Declaration of Competing Interest}
The authors declare no competing financial interests or personal
relationships that could have influenced the work reported in this paper.

\section*{Acknowledgments}
This research did not receive any specific grant from funding agencies
in the public, commercial, or not-for-profit sectors.

\section*{CRediT authorship contribution statement}
\textbf{Md Mahamudur Rahaman Shamim:} Conceptualization, Software,
Formal analysis, Writing - original draft.
\textbf{Md. Nuruzzaman:} Investigation, Data curation, Validation.
\textbf{Zannatul Ferdus:} Investigation, Visualization.
\textbf{Md Rajib Ahmed:} Validation, Resources.
\textbf{Abieer Nwshad Anwar:} Formal analysis, Writing - review \& editing.
\textbf{Mohammad Tooneer:} Resources, Writing - review \& editing.
\textbf{Johir Uddin Khan:} Writing - review \& editing.
\textbf{Khalid Hossen:} Supervision, Methodology.

\section*{Declaration of Generative AI and AI-assisted technologies in the writing process}
During the preparation of this work, the author(s) used Claude (Anthropic)
in order to improve language clarity and readability. After using this
tool/service, the author(s) reviewed and edited the content as needed and
take full responsibility for the content of the published article.

\bibliographystyle{elsarticle-num}
\bibliography{references}

\end{document}